\documentclass[sigconf]{article}
\usepackage{ijcai25}

\usepackage{times}
\usepackage{soul}
\usepackage{url}
\usepackage[hidelinks]{hyperref}
\usepackage[utf8]{inputenc}
\usepackage[small]{caption}
\usepackage{graphicx}
\usepackage{amsmath}
\usepackage{amsthm}
\usepackage{booktabs}
\usepackage{algorithmic}
\usepackage[switch]{lineno}
\usepackage{natbib}

\usepackage{multirow}
\usepackage{makecell}
\usepackage{newfloat}
\usepackage{listings}
\usepackage{xspace}
\usepackage{bibentry}
\usepackage{amssymb}
\usepackage{mathrsfs}
\usepackage[ruled]{algorithm2e}  

\title{Enhancing Generalization in Chain of Thought Reasoning for Smaller Models}

\author{
    Maxwell J. Yin\thanks{Equal contribution.}$^1$,
    Dingyi Jiang\footnotemark[1]$^1$,
    Yongbing Chen$^2$,
    Boyu Wang\thanks{Co-corresponding Author.}$^1$,
    Charles Ling\footnotemark[2]$^1$\\
    \affiliations
    $^1$Western University, London, ON, Canada\\
    $^2$Wenzhou Academy of Agricultural Sciences, Wenzhou, China\\
    \emails
    jyin97@uwo.ca, djiang94@uwo.ca, chenyongbing@wzvcst.edu.cn, bwang@csd.uwo.ca, charles.ling@uwo.ca
}

\begin{document}

\maketitle

\begin{abstract}
    Chain-of-Thought (CoT) reasoning in smaller language models is a challenging natural language process problem yet highly desirable in many real-life applications. Existing CoT knowledge distillation methods often suffer from overly conservative memorization in smaller LLMs, leading to low generalization confidence. As fully preserving the CoT ability of teacher model is impossible, we hypothesize that adversarial CoT fine-tuning is crucial for developing smaller LLM with robust CoT generalization. To this end, we propose \textit{PRompt-Assisted Domain-Adversarial fine-tuning} (PRADA), a principled fine-tuning framework that integrates diverse CoT domains. Specifically, PRADA pioneers two CoT improvements in smaller LLM: (1) Recovering the domain-invariant feature insight which typically lost during distillation with domain adversarial fine-tuning; (2) Enhancing the domain adaptability of CoT prompt engineering by employing domain-adversarial approaches. We theoretically demonstrate the effectiveness of our approach and empirically show that it significantly outperforms the state of the arts in a wide range of tasks. Moreover, our empirical findings reveal that the smaller LLM, when leveraging PRADA, aligns closely with domain knowledge, thereby improving the explainability of our approach.
\end{abstract}

\section{Introduction}

\begin{figure}[t]
\centering
\includegraphics[width=7.5cm]{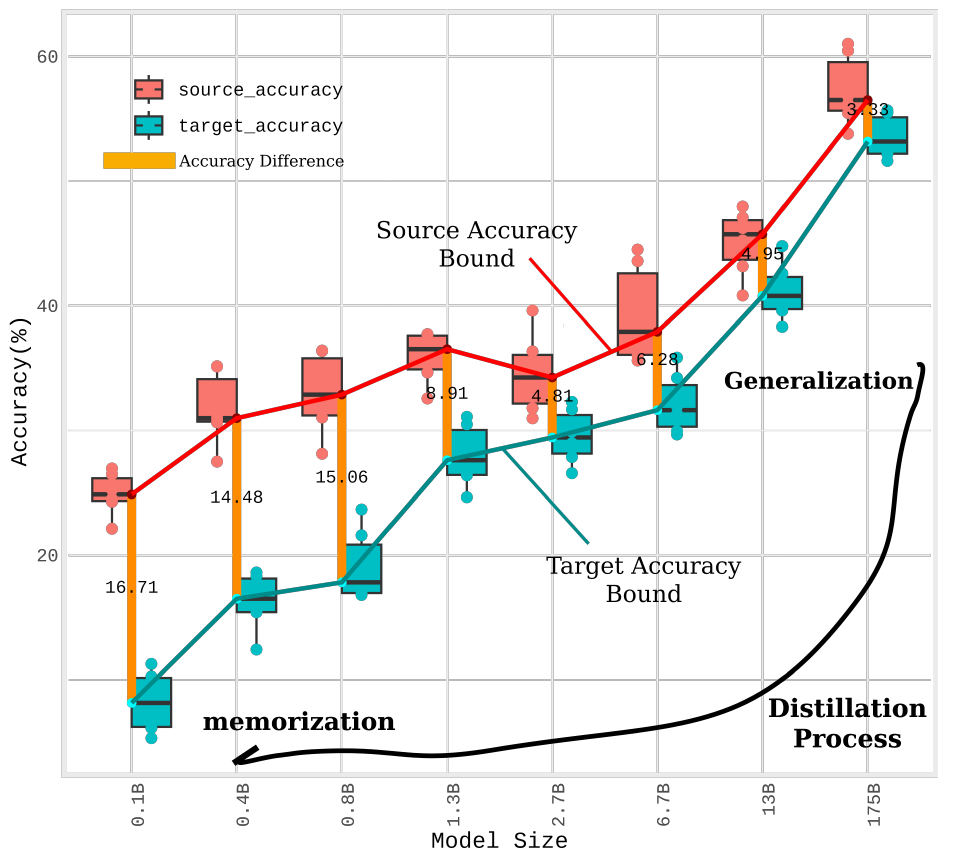} 
\caption{Transition from memorization to generalization, for GPT-3 families fine-tuned on four different domains (datasets detailed in \ref{Experiments}. The source accuracy is tested on the original domain, and the target accuracy is tested on different domains.}
\label{fig2}
\end{figure}

\begin{figure*}[t]
\centering
\includegraphics[width=16cm]{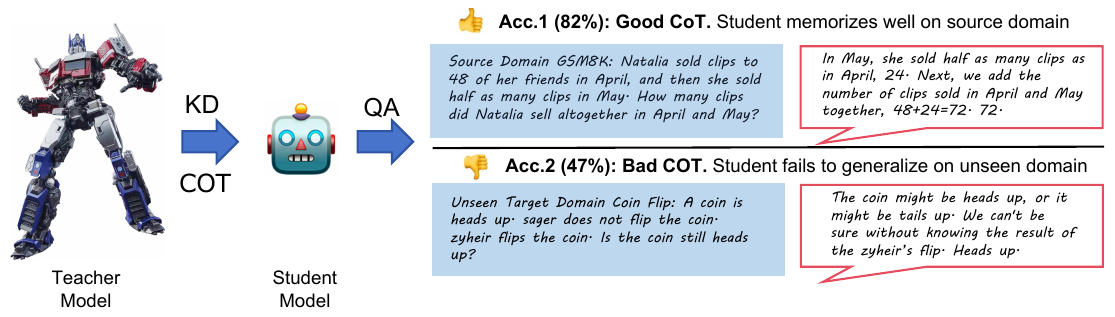}
\caption{Case study of model CoT reasoning ability decay in the knowledge distillation process from GPT-3 175B to GPT-3 6.7B.The reduction in model parameters causes the student model's CoT generalization ability to diminish, shifting towards memorization. Consequently, the student model exhibits bad CoT generalization on unseen data. }
\label{fig1}
\end{figure*}

In recent years, Large Language Models (LLMs) have witnessed significant advancements. To solve complex tasks, prompt-based CoT reasoning \citep{wei_chain--thought_2023} has emerged as a powerful tool to elicit advanced reasoning ability for LLM. Based on this method, LLMs smoothly achieved more cognitive and logical reasoning paths and higher answer correctness by requiring LLMs to perform CoT reasoning. For example, \citet{kojima_large_2023} found that LLMs can be prompted to generate a series of intermediate CoT reasoning steps by simply asking the model to think step by step. 

Whereas the emergence of CoT reasoning is based on extremely large models that usually require more than 100B parameters \citep{wei_chain--thought_2023}, the knowledge distillation of CoT reasoning models \citep{li_explanations_2022} is being used to reduce the number of parameters in LLM. Consequently, the CoT knowledge distillation enables smaller models to imitate the larger model's logical reasoning steps effectively, rather than generating simple and unreliable question-answer pairs \citep{shridhar_distilling_2023,ho_large_2023, fu_specializing_2023}.

However, a major drawback of CoT knowledge distillation is the shift of model ability from generalization to memorization (as shown in Figure 1), as the number of parameters of the student model is drastically reduced compared to the teacher model \citep{rabin_memorization_2023}. In the knowledge distillation setting, the smaller student model often memorizes the pretrained CoT paths \citep{fu_specializing_2023, antoniades_generalization_2024} instead of maintaining the larger teacher model's CoT generalization ability. As shown in Figure 2, the parameter-reduced student model performs CoT reasoning well on the well-informed source domain but fails to generalize to the new, unseen domain. 

To address this, we propose a CoT distillation framework called \textit{PRompt-Assisted Domain-Adversarial fine-tuning} (PRADA): (1) A large teacher model is prompted to generate various CoT reasoning responses. (2) prompt learning layers are incorporated into the smaller model to facilitate the acquisition of domain-agnostic knowledge. (3) Domain-adversarial fine-tuning is applied to both source CoT domain and target domain, enabling the model to learn invariant domain features and mitigate the effects of domain shifts.

We conduct extensive empirical evaluations using 12 datasets in various domains. Our results demonstrate that PRADA substantially outperforms former CoT knowledge distillation methods in cross-domain scenarios. We find that PRADA effectively enhances the cross-domain generalization versatility of smaller models while retaining the CoT reasoning capabilities of the teacher model, which previously relied on \textit{hundreds of billions of} parameters \citep{kojima_large_2023, wei_chain--thought_2023}.

Our contributions can be summarized as:

\begin{itemize}
    \item {Enhancing the domain discrimination capability of LLMs through an adversarial fine-tuning framework.} 
    \item {Improving the domain adaptability of CoT prompt tuning via adversarial learning.}
    \item {Addressing generalization degradation in the knowledge distillation process for Chain-of-Thought reasoning}

\end{itemize}

\section{Related Work}

\paragraph{Knowledge Distillation Based on Chain-of-Thought Reasoning} In the evolving realm of knowledge distillation (KD) and chain-of-thought (CoT) reasoning, the MT-COT framework \citep{li_explanations_2022} stands out by utilizing COT explanations generated by LLMs to bolster the training of compact reasoning engines. Building on this foundation, the Specialized model\citep{fu_specializing_2023} extracts CoT reasoning paths from large teacher models to enhance out-of-distribution generalization. This specialized extraction is crucial for improving the adaptability and accuracy of models when encountering unfamiliar data distributions. Further advancing the field, Fine-tune CoT\citep{ho_large_2023} introduces a technique where multiple COT reasoning solutions are generated from LLMs through random sampling. This method addresses the challenge of limited training data diversity, which often hampers the generalization performance of student models. Innovatively, the SOCRATIC CoT\citep{shridhar_distilling_2023} splits the training into two distilled models: a problem decomposer and a subproblem solver. This approach mitigates the cognitive load on student models, allowing for more effective learning and CoT reasoning. 

\paragraph{Domain Adversarial Neural Network}
Our research is closely aligned with Adversarial Training (AT) \citep{goodfellow_explaining_2015}, a prevalent machine-learning technique for enhancing model robustness. Building on this foundation, the Domain-Adversarial Neural Network (DANN) \citep{ganin_domain-adversarial_2016} is designed to improve the model's generalization ability. In the field of Natural Language Processing (NLP), the standard approach involves applying adversarial perturbations to word embeddings \citep{miyato_adversarial_2021}. This technique has been demonstrated to significantly enhance model generalization ability when used for fine-tuning across various downstream tasks, as evidenced by methods such as FreeLB\citep{zhu_freelb_2020}, SMART\citep{jiang_smart_2020}, InfoBERT\citep{wang_infobert_2021}, and CreAT\citep{wu_toward_2023}. Additionally, ALUM\citep{liu_adversarial_2020} provides empirical evidence that adversarial training can yield substantial pre-training benefits. Our work advances fine-tuning processes for pre-trained language models (PrLMs) \citep{devlin_bert_2019, liu_adversarial_2020, raffel_exploring_2023, he_deberta_2021}. It is independent of the prevailing pre-training paradigms, such as Masked Language Modeling (MLM) \citep{devlin_bert_2019}, Replacement Token Detection (RTD)\citep{clark_electra_2020}, Permuted Language Modeling (PLM)\citep{cui_pert_2022}, and multiple-objective approaches\citep{wu_forging_2022}.

\begin{figure*}[ht]
\centering
\includegraphics[width=15cm]{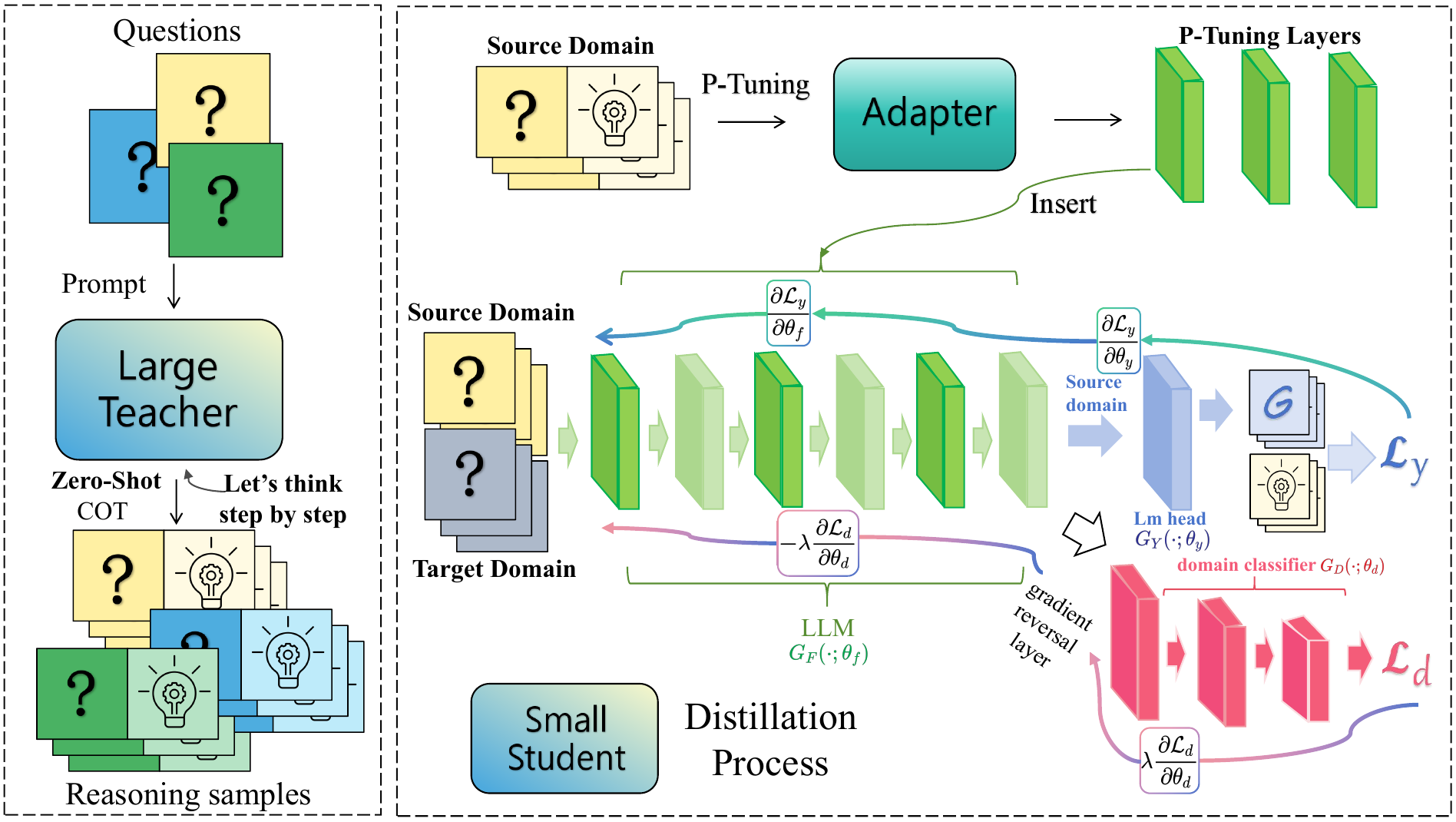} 
\caption{Illustration of our method Prompt-Assisted Domain-Adversarial fine-tuning (PRADA). \textbf{Firstly}, the teacher model is prompted to generate diverse CoT reasoning responses using a Zero-Shot-CoT approach. \textbf{Secondly}, an adapter is inserted into the student LLM for P-Tuning, which refines the domain-agnostic knowledge of the source domain. \textbf{Thirdly}, the knowledge distillation process incorporates data from both the labeled source domain and the unlabeled target domain. The proposed architecture includes an LLM (green) and a language modeling head (blue), which together form a standard feed-forward architecture. Unsupervised domain adversarial fine-tuning is achieved by adding a domain classifier (red).}
\label{fig3}
\end{figure*}

\paragraph{Prompt Learning} The field of prompt learning has witnessed significant advancements, providing innovative approaches to enhance Natural Language Understanding (NLU) tasks. "Prefix Tuning," introduced by \citep{li_prefix-tuning_2021}, improves model comprehension by appending prefixes to input sequences. The "WARP" method by \citep{hambardzumyan_warp_2021} and "P-tuning" by \citep{liu_gpt_2023} modify language model outputs and input sequences, respectively, achieving robustness comparable to traditional fine-tuning while employing fewer task-specific parameters. Domain Adaptation via Prompt Learning (DAPL)\citep{ge_domain_2022} leverages pretrained vision-language models by optimizing a minimal number of parameters to embed domain-specific information into prompts. This method enables the classifier to dynamically adapt to each domain. Prompt-Assisted Self-Adaptive Learning (PASAL)\citep{yin_source-free_2024} utilizes prompts to facilitate the acquisition of domain-agnostic knowledge, effectively addressing domain shifts.

\section{Methodology}

\subsection{Problem Definition}

Given a set of source questions $\mathcal{D}_s = \{(q_i^s)\}_{i=1}^{N_s}$ and a set of target questions $\mathcal{D}_t = \{(q_i^t)\}_{i=1}^{N_t}$, we adopt a model trained from a source domain to a target domain simultaneously. Here, $N_s$ and $N_t$ denote the sizes of the source domain dataset $\mathcal{D}_s$ and the target domain dataset $\mathcal{D}_t$, respectively. Let $G_F (\cdot; \theta_f)$ be a pretrained language model with a hidden state size of $H$ and a vocabulary size of $|\mathcal{V}|$. The input embeddings are given through the pretrained embedding layer, where $e \in \mathbb{R}^{|\mathcal{V}| \times H}$. In the following, we characterize the general PRADA architecture in three steps, which are also visualized in Figure 3.

\subsection{Teacher: Diverse Reasoning Generation}

First, we leverage a large teacher model to generate a variety of CoT reasoning responses for a given task. To enhance usage flexibility and minimize teacher inference costs, we employ the task-agnostic Zero-Shot-CoT prompting method \citep{kojima_large_2023} on teacher models. Consider a standard sample $ S_i $ comprising a question $ q_i $ and its correct answer $ a_i $. By using Zero-Shot-CoT, we prompt the teacher model to generate a reasoning rationale, $ \hat{r}_i $, to address question $ q_i $ and provide a final answer prediction $ \hat{a}_i $. The result text sequence includes the question and generated content:“ Q: $\langle q_i \rangle$. A: Let’s think step by step. $\langle  \hat{r}_i  \rangle$ Therefore, the answer is $\langle  \hat{a}_i  \rangle$." To enhance reasoning ability, we leverage the effort of diverse CoT reasoning. For a given sample $S_i$, we employ a stochastic sampling strategy, specifically temperature sampling with a large $t$, to generate $D$ distinct outputs $\{(\hat{r}_{ij}, \hat{a}_{ij})\}_{j=1}^{D}$. We refer to $D$ as the degree of reasoning diversity.

Next, we filter the generated samples and reformat them into prompt-completion pairs. For filtering, we compare the teacher model’s final prediction $ \hat{a}_i $ with the ground truth answer $ a_i $ \citep{huang_large_2022, zelikman_star_2022}. For all instances where $ \hat{a}_i = a_i $, we repackage $ (S_i, \hat{r}_i, \hat{a}_i) $ into a reasoning sample $ S_i' = (q_i, c_i) $, a question-completion pair. Specifically, $ q_i $ and $ c_i $ take the forms of “$\langle q_i \rangle$ \#\#\#” and “$\langle \hat{r}_i \rangle --\rangle \langle  a_i  \rangle$ END”. 

\subsection{Student: Prompt Learning }

We start with a source domain $\mathcal{D}_s$ comprising CoT reasoning samples $\{(q_i, c_i)\}_{i=1}^{N_s}$, where $q_i$ represents the question sequence and $c_i$ denotes the corresponding completion. Next, we deploy the P-Tuning adapter on the student model, which employs continuous prompt embeddings to enhance stability and improve prompting performance. Let $[P]$ represent the continuous prompt embedding. The prompt template for P-Tuning is defined as follows:

\begin{equation} \label{eq2}
\begin{split}
T_i = \{P_{[0:j]}, q_i, P_{[(j+1):k]}, c_i, P_{[(k+1):l]}\}
\end{split}
\end{equation}

P-Tuning leverages an additional embedding function $f : [P_i] \rightarrow h_i$ to map the template to:

\begin{small}  
\[
\{h_0, \ldots, h_j, e(q_i), h_{j+1}, \ldots, h_k, e(c_i), h_{k+1}, \ldots, h_l\}
\]
\end{small}
During the prompt learning, the embeddings $\{P_i\}_{i=1}^k$ are updated to optimize a task loss function $\mathcal{L}_p$ using the template. These multiple P-tuning layers are trained as an adapter on the source domain, which employs deep prompt learning that involves a lot of P-tuning layers to improve and stabilize prompting \citep{liu_p-tuning_2022}. Then, we insert the adapter into the student LLM. When the student model is fine-tuning or inferencing, the prompts in different P-tuning layers are added as prefix tokens, providing the student LLM with more tunable domain-agnostic parameters of the source domain data \citep{yin_source-free_2024}.

\subsection{Student: Domain Adversarial Fine-Tuning}

Let $G_F (\cdot; \theta_f)$ represent the student LLM, with parameters $\theta_f$. Then, let $G_Y(\cdot; \theta_y)$ denote the natural language modeling head, with parameters $\theta_y$, while $G_D(\cdot; \theta_d)$ corresponds to the domain classifier of the PRADA architecture, with parameters $\theta_d$. The input source domain data $\{(q_i, c_i)\}_{i=0}^{N_s}$ and target domain data $\{(q_i)\}_{i=0}^{N_t}$ are transformed into $\{(q_i, c_i, d_s)\}_{i=0}^{N_s}$ and $\{(q_i, d_t)\}_{i=0}^{N_t}$, respectively, where $d_s$ and $d_t$ are unique symbols distinguishing the domains. We denote the prediction loss and the domain loss, respectively, :

\begin{equation} \label{eq3}
L^i_y(\theta_f, \theta_y) = L_y (G_y(G_f (q_i; \theta_f); \theta_y), c_i)
\end{equation}
\begin{equation} \label{eq4}
L^i_d(\theta_f, \theta_d) = L_d (G_d(G_f (q_i; \theta_f); \theta_d), d_i)
\end{equation}

Then, training PRADA parallels the single-layer case and consists of optimizing:

\begin{equation} \label{eq5}
\begin{split}
&E(\theta_f, \theta_y, \theta_d) = \frac{1}{N_s} \sum_{i=1}^n L^i_y(\theta_f, \theta_y)
\\ & - \lambda ( \frac{1}{N_s} \sum_{i=1}^{N_s} L^i_d(\theta_f, \theta_d)+\frac{1}{N_t} \sum_{i=1}^{N_t} L^i_d(\theta_f, \theta_d) ),
\end{split}
\end{equation}

A saddle point can be found as a stationary point of the following gradient updates\citep{ganin_domain-adversarial_2016}:

\begin{equation} \label{eq6}
\begin{split}
\theta_f \leftarrow \theta_f - \mu \left( \frac{\partial L^i_y}{\partial \theta_f} - \lambda \frac{\partial L^i_d}{\partial \theta_f} \right),
\end{split}
\end{equation}

\begin{equation} \label{eq7}
\begin{split}
\theta_y \leftarrow \theta_y - \mu \frac{\partial L^i_y}{\partial \theta_y},
\end{split}
\end{equation}

\begin{equation} \label{eq8}
\begin{split}
\theta_d \leftarrow \theta_d - \mu \lambda \frac{\partial L^i_d}{\partial \theta_d},
\end{split}
\end{equation}

\begin{algorithm}
    \caption{PRADA Algorithm}
    \KwIn{Source questions $\mathcal{D}_s$, Target questions $\mathcal{D}_t$, Teacher LLM $\boldsymbol{F}$, Domain Classifier $\boldsymbol{D}$}
    \KwOut{Learned student LLM $\boldsymbol{f}$}
    
    \underline{\emph{Step 1: Teacher - Diverse CoT Generation}} \\
    \For {each $q_i \in \mathcal{D}_s$}{
        Obtain the CoT reasoning $c_i \gets \boldsymbol{F}(q_i)$\;
    }
    Obtain the Query-CoT Pairs $\mathcal{C}_s \gets \{(q_i, c_i)\ |\ q_i \in \mathcal{D}_s\}$\;
    
    \underline{\emph{Step 2: Student - Domain Adversarial Fine-Tuning}} \\
    Initialize Adapter $\boldsymbol{f}$ using $\boldsymbol{P}$-\texttt{Tuning} on $\mathcal{C}_s$\;
    
    \While{\emph{not converged}}{
        Sample $(q^s_i, c^s_i) \in \mathcal{C}_s$, $q^t_i \in \mathcal{D}_t$\;
        
        Compute task-specific loss $L^i_y$ using $(q^s_i, c^s_i)$ via Eq.~(2)\;
        
        Compute domain adversarial loss $L^i_d$ using $q^s_i$ and $q^t_i$ via Eq.~(3)\;
        
        Update $\boldsymbol{f}$ and $\boldsymbol{D}$ via Eq.~(10)\;
    }
\end{algorithm}

where $\mu$ is the learning rate and $\lambda$ is the domain classifier loss weight. We use stochastic estimates of these gradients by sampling examples from the dataset. The only difference is that in Equation (3), the gradients from the LLM and domain predictors are subtracted, instead of being summed. This distinction is crucial, as otherwise, SGD would attempt to make features dissimilar across domains in order to minimize the domain classification loss. Fortunately, this can be achieved by introducing a special gradient reversal layer (GRL)\citep{ganin_domain-adversarial_2016}, defined as follows:

\begin{equation} \label{eq9}
\begin{split}
R(q) = q,
\end{split}
\end{equation}

\begin{equation} \label{eq10}
\begin{split}
\frac{dR}{dq} = -I,
\end{split}
\end{equation}

where $I$ is an identity matrix. We can then define the objective “pseudo-function” of $(\theta_f, \theta_y, \theta_d)$ that is being optimized by the stochastic gradient descent within our method:

\begin{equation} \label{eq11}
\begin{split}
&\tilde{E}(\theta_f, \theta_y, \theta_d) = \frac{1}{N_s} \sum_{i=1}^{N_s} L_y (G_y(G_f (q_i; \theta_f); \theta_y), c_i)\\
&-\lambda ( \frac{1}{N_s} \sum_{i=1}^{N_s} L_d (G_d(R(G_f (q_i; \theta_f)); \theta_d), d_i ) \\&+ \frac{1}{N_t} \sum_{i=1}^{N_t} L_d (G_d(R(G_f (q_i; \theta_f)); \theta_d), d_i ) ).
\end{split}
\end{equation}

\begin{table*}[ht]
    \centering
    \small
    \tabcolsep=0.15cm
    \renewcommand\arraystretch{0.3}

   \begin{tabular}{lccccccc}
    \toprule[1.1pt]
    & \multicolumn{7}{c}{\textbf{Comparison Experiments for PRADA on Arithmetic Reasoning Problems}} \\
    \cmidrule(lr){2-8} 
    \textbf{Models} & \textbf{\makecell[c]{\#Source\\Domain}} & \textbf{SVAMP} & \textbf{MultiArith} & \textbf{AQUA} & \textbf{Last Letter} & \textbf{Strategy QA} & \textbf{Tracking Shuffled} \\
    & & \textbf{Acc.}  & \textbf{Acc.}  & \textbf{Acc.}  & \textbf{Acc.}  & \textbf{Acc.}  & \textbf{Acc.} \\
    \midrule[1.1pt]
    Specialized, T5 & \multirow{5}{*}[-2ex]{GSM8K} & 24.21 & 27.5 & 14.56 & 19.49 & 58.7 & 27.6 \\
    MT-CoT, T5 &  & 22.8 & 15.45 & 19.82 & 16.94 & 79.35 & 32.87 \\
    PRADA, T5 &  & 26.93  & 28.15 & 19.85  & 67.85 & 81.56 & 38.36 \\
    Fine-Tune-CoT, Llama3 &  & 52.85 & 52.05 & 52.13 & \textbf{41.67} & 88.53 & 79.8 \\
    PRADA, Llama3 &  & \textbf{53.07} & \textbf{53.11} & \textbf{57.47} & 38.2 & \textbf{91.42} & \textbf{89.25} \\
    \midrule[1.1pt]
    Specialized, T5 & \multirow{5}{*}[-2ex]{Coin Flip} & 32.12 & 25.50 & 14.62 & 61.4 & 63.72 & 28.62 \\
    MT-CoT, T5 & & 21.81 & 24.47 & 18.85& 66.91 & 78.13& 21.38\\
    PRADA, T5 & & 35.68 & 32.84 & 16.48 & 81.24 & 84.68 & 30.97 \\
    Fine-Tune-CoT, Llama3 & & \textbf{42.88} & 32.01 & 31.13 & 90.67& 87.53& 88.8   \\
    PRADA, Llama3 & & 42.63 & \textbf{34.09} & \textbf{38.00} & \textbf{92.10} & \textbf{90.44} & \textbf{96.99} \\
    \midrule[1.1pt]
    Specialized, T5 & \multirow{5}{*}[-2ex]{\scriptsize CommonSense QA} & 16.23& 22.57 & 21.65 & 30.48 & 33.71& 29.62\\
    MT-CoT, T5 & & 12.12 & 23.74 & 27.81 & 25.99 & 27.32 & 30.85  \\
    PRADA, T5 & & \textbf{33.86} & 35.59 & 27.56 & 59.48 & 38.64 & 53.78 \\
    Fine-Tune-CoT, Llama3 & & 31.81 & 34.01 & 30.13 & 90.67 & 47.53& 78.18 \\
    PRADA, Llama3 & & 27.05 & \textbf{42.39} & \textbf{39.14} & \textbf{94.92} & \textbf{50.86} & \textbf{79.52} \\
    \midrule[1.1pt]
    Specialized, T5 & \multirow{5}{*}[-2ex]{\scriptsize Date Understanding} & 23.23 & 32.25 & 21.06 & 61.42 & 70.17& 57.36 \\
    MT-CoT, T5 & & 20.81 & 33.41 & 17.83 & 55.97 & 77.33 & 60.88\\
    PRADA, T5 & & 25.19 & 34.17 & 20.03 & 61.18 & \textbf{94.72} & 72.49 \\
    Fine-Tune-CoT, Llama3 & & 31.89 & 34.03 & 40.13 & 90.67 & 77.53 & 78.84 \\
    PRADA, Llama3 & & \textbf{33.96} & \textbf{37.2} & \textbf{41.08} & \textbf{92.1} & 76.11 & \textbf{84.58} \\
    \bottomrule[1.1pt]
\end{tabular}

    \caption{Comparison of PRADA(ours) with the related methods for reasoning tasks. Each row represents the accuracy (\%) of a compared method adapting between the source and target domain. }
    \label{results1}
\end{table*}

Notably, the overall student parameter update includes the update of P-tuning layers. To maintain the effectiveness of P-tuning layers, specifically the domain-agnostic learning ability of P-tuning layers learned on the source domain, we use a learning rate difference strategy between the training process of the adapter and the LLM. In the prompt learning process, we utilize the learning rate in the order of $e^{-3}-e^{-4}$. However, the overall student model fine-tuning requires the learning rate in the order of $e^{-5}-e^{-6}$. This learning rate difference ensures that the previous prompt learning effect will not be significantly affected while domain-adversarial fine-tuning.

\section{Experiments}\label{Experiments}

\paragraph{Tasks and Datasets} We assessed the effectiveness of our approach using 12 datasets that span four categories of complex reasoning, in line with the framework established by \citet{kojima_large_2023}. These categories are:
\begin{itemize}
    \item Arithmetic Reasoning: SingleEq, AddSub, MultiArith, GSM8K, and SVAMP, which test various aspects of mathematical problem-solving and numerical reasoning.
    \item Symbolic Reasoning: Last Letter Concatenation and Coin Flip are used to evaluate reasoning that involves symbolic manipulation and pattern recognition.
    \item Commonsense Reasoning: CommonSenseQA and StrategyQA, which test the ability to apply everyday knowledge and reasoning strategies to answer questions.
    \item Other Reasoning:Date Understanding and Tracking Shuffled Objects, focusing on logical reasoning skills and the ability to track and understand sequences or temporal data.
\end{itemize}
For each dataset, we follow \citep{ ho_large_2023}, using the teacher model's various CoT reasoning for downstream student model fine-tuning. 

\paragraph{Teacher/Student Models} For the teacher models, we utilize the GPT-3 175B \citep{brown_language_2020}, accessible via the OpenAI API. Specifically, the teacher model used for PRADA is text-davinci-002, which is based on Instruct-GPT 175B \citep{ouyang_training_2022}. The experiments involving student models are categorized into two types: Encoder-Decoder Architecture and Decoder-Only Architecture. For the decoder-only architecture, we choose the newly updated Llama3 model\citep{dubey_llama_2024} for Lora fine-tuning \citep{hu_lora_2021}. By the way, we originally wanted to choose GPT-3 and GPT-2 \{GPT 2-small-125M, GPT 2-median-335M, GPT 2-large-774M\} as models, but because of the black-box nature of GPT 3 and the lack of mathematical ability of GPT 2, we discarded GPT families for the decoder-only model. For the encoder-decoder architecture, we incorporate Flan-T5-Large \citep{chung_scaling_2022}, an instruction-tuned version of T5, to evaluate the impact of the PRADA on the encoder-decoder model. These student models are 25 to 250 times smaller than the teacher model, making them more practical for real-world deployment. 

\paragraph{Baselines}
We provide a comparison of PRADA (ours) with three CoT knowledge distillation methods:
\begin{itemize}

    \item \textbf{Specialized LLM} \citep{fu_specializing_2023}: This baseline extracts specialized reasoning paths, or Chains of Thought, from a large teacher model to enhance out-of-distribution generalization ability. By fine-tuning on these specialized CoT paths, the student model is expected to achieve superior performance on tasks requiring complex reasoning and extrapolation beyond the training distribution.

\begin{figure}[ht]
\centering
\includegraphics[width=8.8cm]{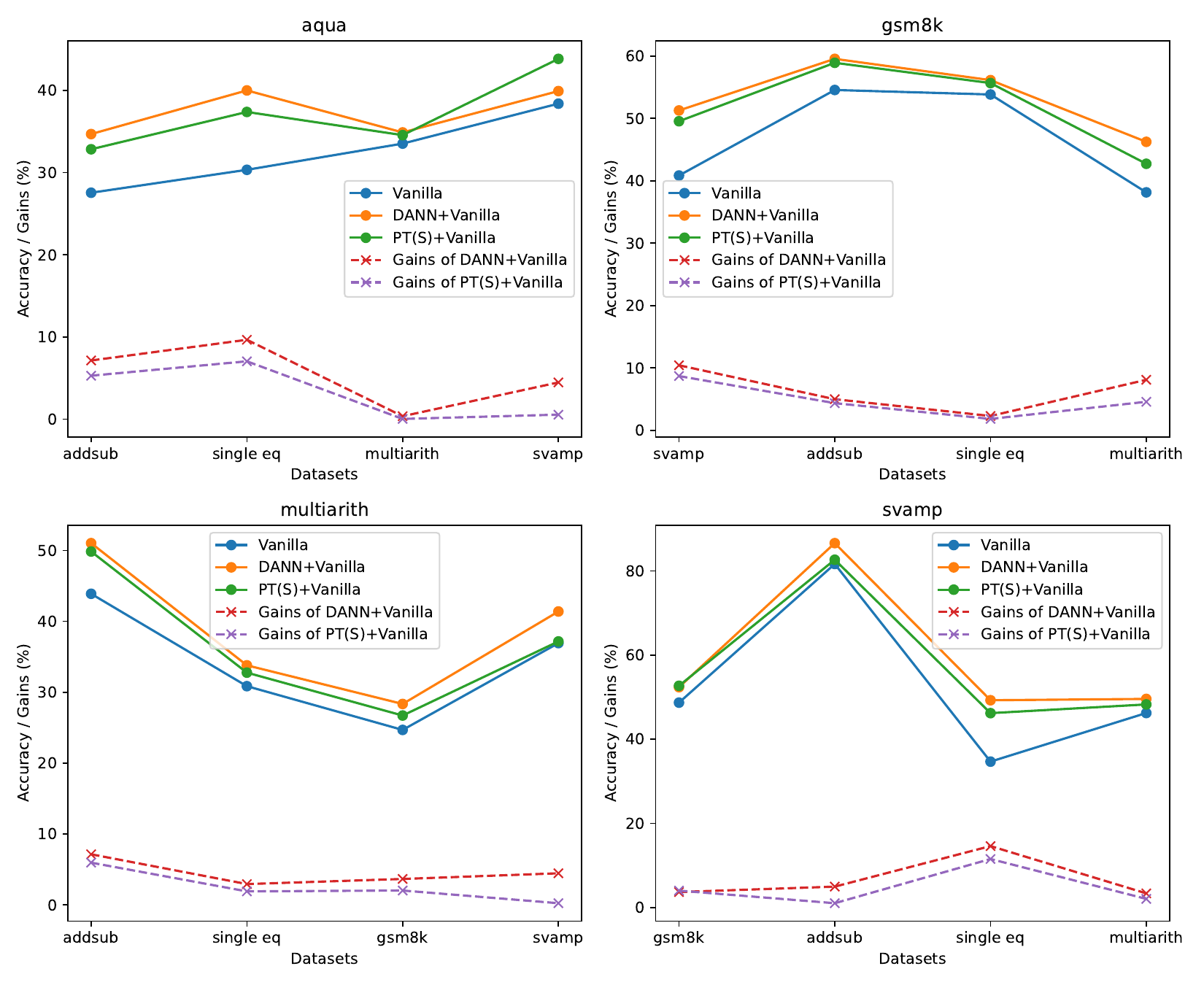} 
\caption{Ablation study (Acc. \%) of Components of PRADA: Vanilla Llama3 versus Enhanced with Additional Components.}
\label{fig4}
\end{figure}

    \item \textbf{Fine-tune CoT} \citep{ho_large_2023}: This method involves randomly sampling multiple reasoning solutions generated by LLMs. The augmentation of training data with diverse reasoning paths facilitates the learning process of the student model, enabling it to develop robust reasoning abilities through exposure to varied inferential strategies.

    \item \textbf{MT-CoT (Multi-task Chain of Thought)} \citep{li_explanations_2022}: This approach leverages explanations generated by large language models to enhance the training of smaller reasoning models. By integrating the interpretative ability of LLMs, MT-COT aims to improve the reasoning performance of compact models through enriched training data.
\end{itemize}

\subsection{Main Results}

We conducted extensive experiments on 12 datasets, comparing our Prompt-Assisted Domain-Adversarial fine-tuning (PRADA) models with several baseline and contemporary models. Specifically, we compared against the pretrained versions of Davinci-002, FlanT5-Large, and Llama3-8B, as well as the fine-tuned models: Fine-tune CoT, Specialized, and MT-COT. Our evaluation metrics were accuracy (Acc.) and the improvement in accuracy ($\Delta$). For a fair comparison, we ensured that all methods followed the same experimental settings. Table 1 shows the performance of all models across multiple datasets, recording the average accuracy of target domain samples across various arithmetic reasoning tasks. Our PRADA-enhanced models consistently achieved the highest accuracy across all datasets. For instance, in the source SVAMP dataset, Llama3-8B + PRADA outperformed its pretrained counterpart by over 10\% in accuracy.

\subsection{Ablation Study} 

\begin{figure}[ht]
\centering

\includegraphics[width=8.7cm]{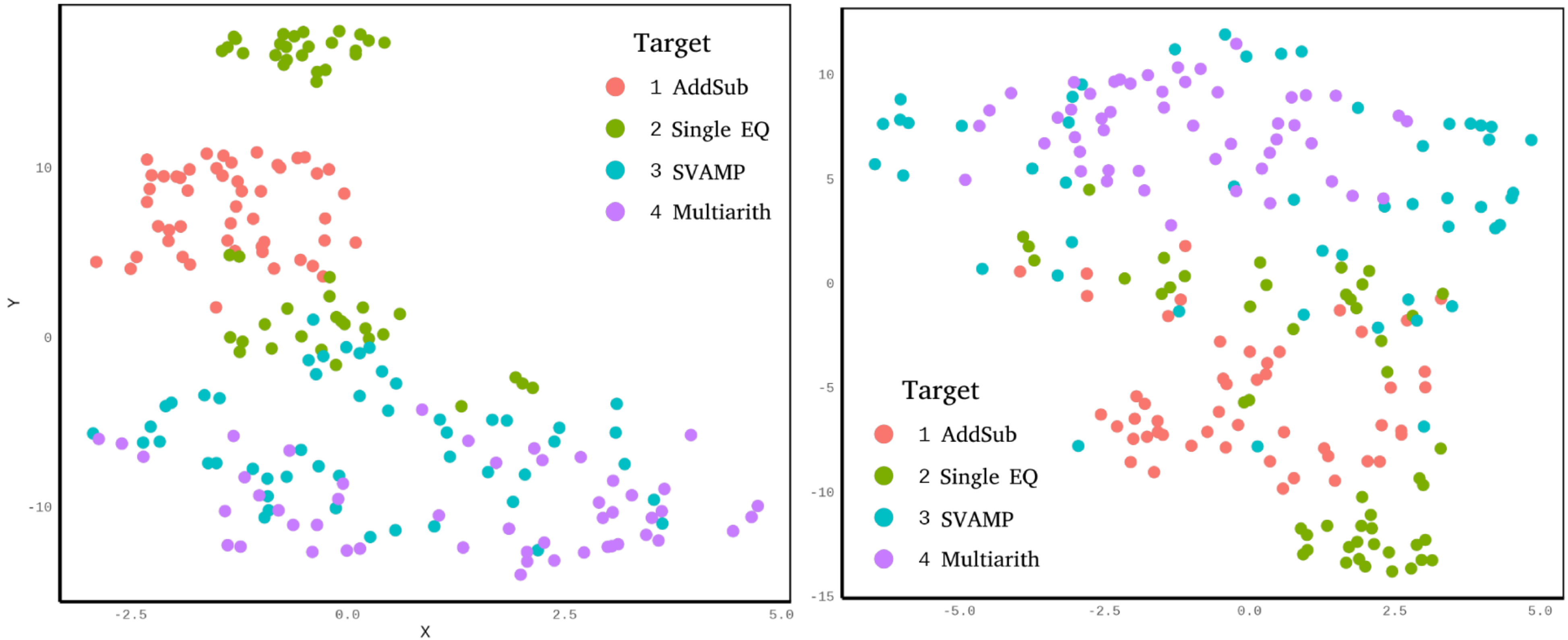} 
\caption{The t-SNE visualization of embeddings of Llama 3-8B using our method PRADA or not on source GSM8K for 4 arithmetic targets.}
\label{fig5}
\end{figure}

\paragraph{Without Domain Classifier.} First, we investigated the impact of the domain classifier by eliminating it from the model. This experiment was designed to explore the domain classifier's role in facilitating effective model generalization ability. A decline in target domain performance following the removal of the domain classifier would suggest that it is vital for distinguishing domain-specific features and thereby maintaining robust performance across different domains. The results of these experiments are visualized in Figure 4. A specific case where the model was trained on GSM8K as the source domain and evaluated across various datasets is presented in Table 2.

\paragraph{Without P-Tuning Layers.} Next, we assessed the role of P-Tuning layers by removing them from the model. The objective here was to determine how crucial these layers are to the system's overall performance. By comparing the results of this modified model against the full version that includes P-Tuning, we aimed to identify any significant drops in performance that would indicate the importance of these layers in enhancing the model generalization ability. The results of these experiments are visualized in Figure 4. A specific case is presented in Table 2.

\begin{figure*}[ht]
	
	\begin{minipage}{0.52\linewidth}
		\vspace{1pt}
		\centerline{\includegraphics[width=\textwidth]{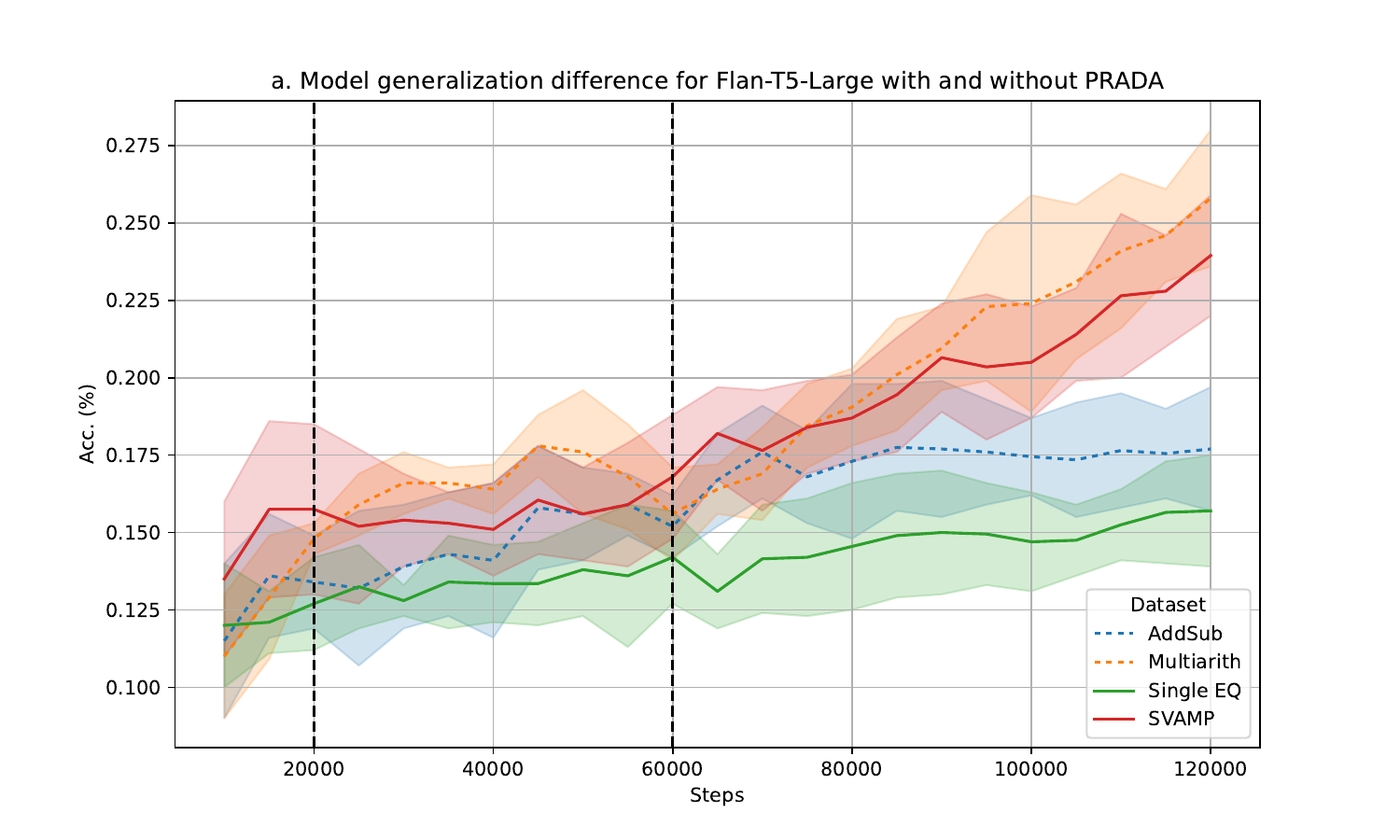}}
	\end{minipage}
	\begin{minipage}{0.52\linewidth}
		\vspace{1pt}
		\centerline{\includegraphics[width=\textwidth]{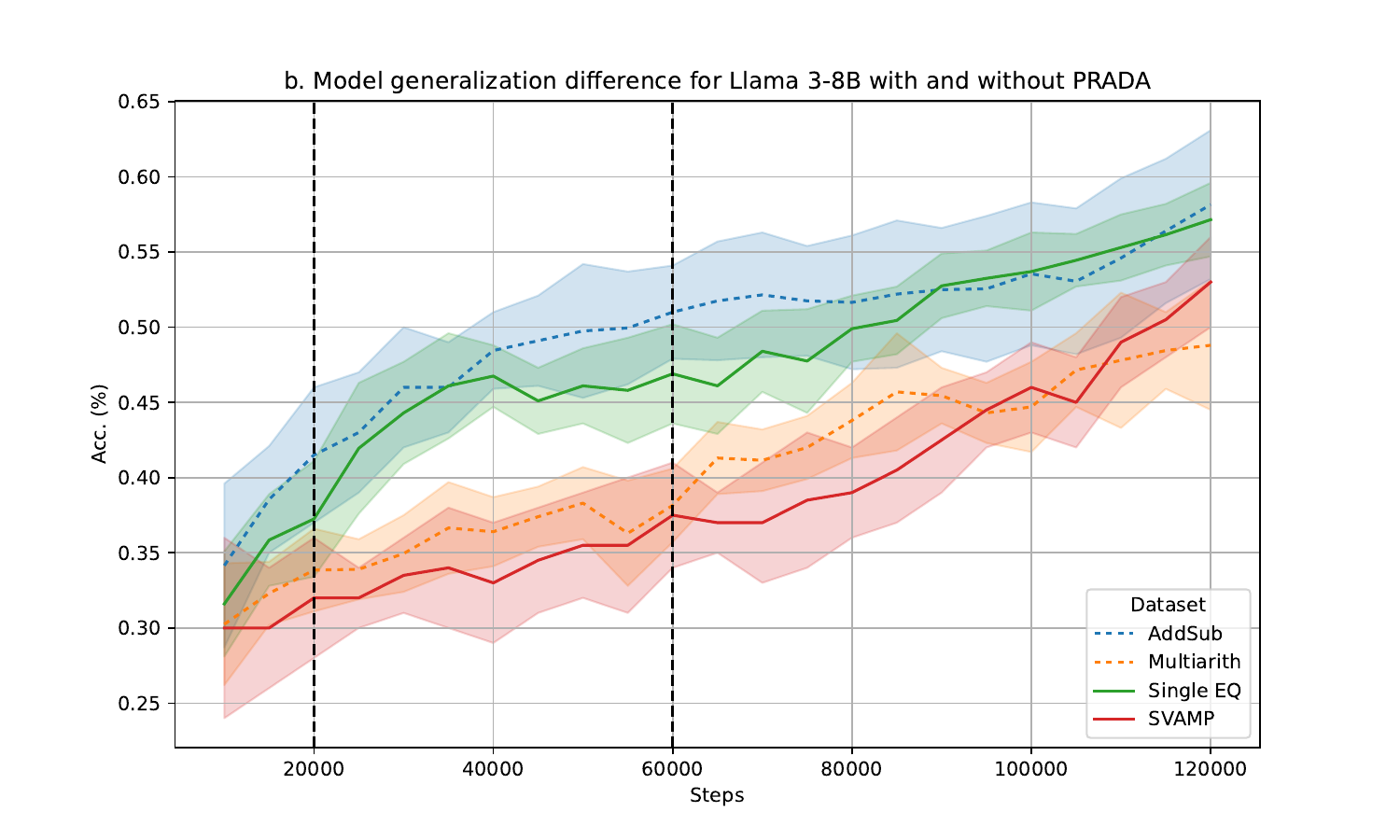}}
	 
	\end{minipage}
 
	\caption{The generalization ability of models with and without PRADA (ours) finetuned on source domain GSM8K. The lower limit of the same color curve is the Vanilla model's accuracy, and the upper limit is the PRADA method's accuracy. }
	\label{fig6}
\end{figure*}

\paragraph{Visualization of embedding with and without PRADA.} The t-SNE visualizations (shown in Figure 5) compare the embeddings produced by LLM with and without the PRADA method across four arithmetic reasoning tasks. On the left of Figure 5, embeddings are shown for the vanilla Llama3 model fine-tuned on GSM8K as the source domain. On the right, embeddings are displayed for Llama3 utilizing PRADA, also trained on GSM8K as the source domain. The visualizations demonstrate the impact of prompt learning in aligning domain-capturing features, with the PRADA approach showing a more generalized clustering of embeddings across different tasks.

\paragraph{Difficulty of Transfer Learning} For arithmetic reasoning datasets, we observe a ranking in terms of the difficulty of transfer with respect to dataset characteristics, which influences the transfer learning performance of PRADA. First, the AQUA dataset involves complex arithmetic tasks, such as solving systems of equations and combinatorics. Secondly, datasets like GSM8K, MultiArith, SVAMP, and AddSub involve more basic arithmetic operations, such as addition, subtraction, multiplication, and division. Based on these preliminary observations, we analyze the results in Table 1 to assess the difficulty within these arithmetic tasks. In particular, we observe that the sets GSM8K and AQUA appear to be too difficult for any small student model, given that the teacher model gets below 65\% accuracy on both. The arithmetic task transfer difficulty decreasing from Left can be ranked as follows: AQUA, GSM8K, Multiarith, SVAMP, Single EQ, Addsub. By the way, this generalization difficulty based on domain-agnostic features is worth studying on other datasets.

\begin{table}[ht]
\centering
\renewcommand\arraystretch{0.25}
\scalebox{1}{
\begin{tabular}{@{}ccc|cccc@{}}
\toprule[1.3pt]
\multicolumn{3}{c|}{\textbf{Components}} & \multicolumn{4}{c}{\textbf{Target Domains}} \\ \midrule
$\mathcal{L}_{y}$ & $\mathcal{L}_{p}$ & $\mathcal{L}_{d}$ & addsub & multiarith & single eq & svamp \\ \midrule
$\checkmark$ &  &  & 54.56 & 44.89 & 54.53 & 44.39 \\
$\checkmark$ & $\checkmark$ &  & 58.92 & 47.88 & 55.84 & 49.84 \\
$\checkmark$ &  & $\checkmark$ & 59.55 & 51.28 & 56.85 & 51.02 \\ 
$\checkmark$ & $\checkmark$ & $\checkmark$ & \textbf{62.68} & \textbf{53.11} & \textbf{59.01} & \textbf{53.07} \\ 
 \bottomrule[1.3pt]
\end{tabular}
}
\caption{Ablation study (Acc. \%) of PRADA based on llama3 without different components are fine-tuned and evaluated from a domain (GSM8K) to the other four arithmetic tasks.}
\end{table}
\paragraph{Generalization ability Analysis} 
In this part, we illustrate the enhancement in generalization ability for two models, Flan-T5-Large and Llama 3-8B, when utilizing the PRADA method compared to the vanilla model counterparts. Figure 6 displays the accuracy difference between models with and without PRADA as the training steps increase. For each dataset (AddSub, MultiArith, SingleEq, SVAMP), the upper limit of the shaded area represents the accuracy achieved by the model with PRADA, while the lower limit represents the accuracy without PRADA. The visualized result shows that PRADA can capture the essence of the generalization visualizations in the context of model generalization evaluation. The area between these curves reflects the improvement in model generalization due to the PRADA method's domain-agnostic feature capturing, highlighting its effectiveness across multiple domains.
\begin{figure}[ht]
\centering
\includegraphics[width=8.7cm]{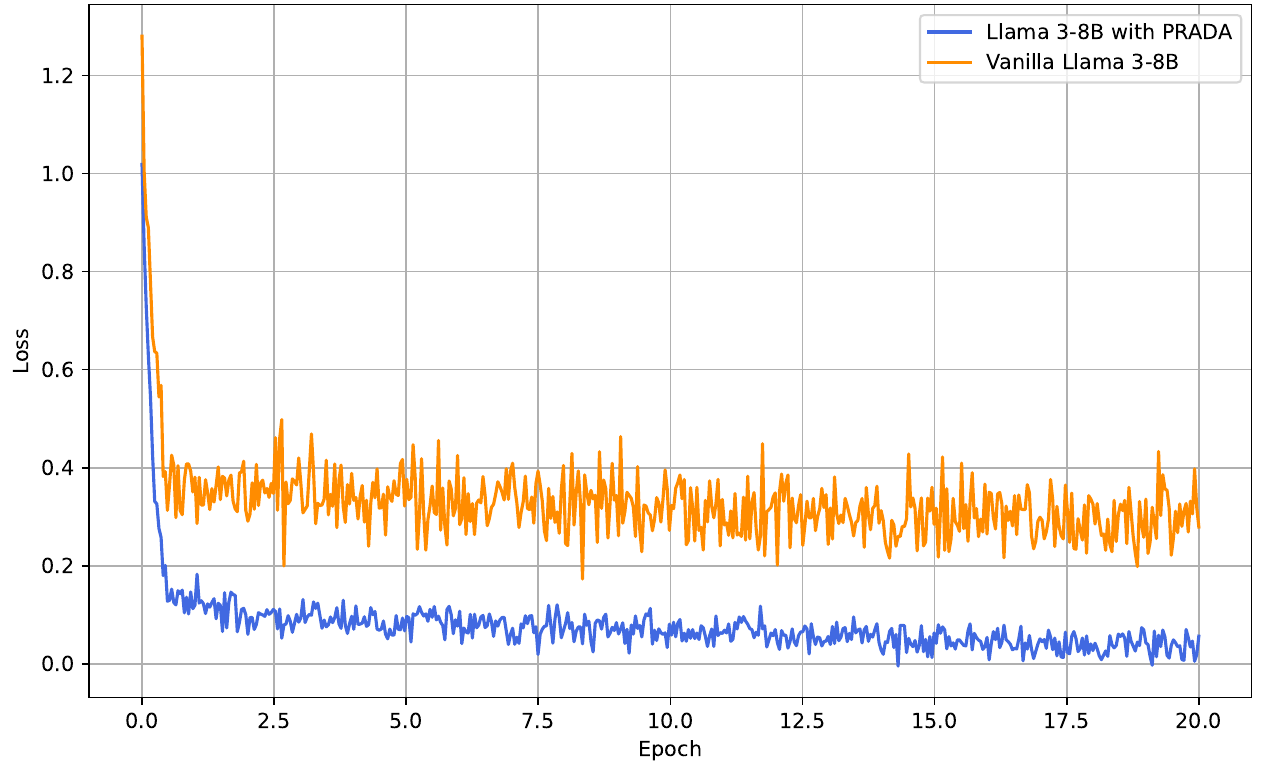} 
\caption{Convergence curve of Llama 3-8B with and without PRADA fine-tuned on GSM8K dataset.}
\label{fig7}
\end{figure}
\paragraph{Convergence Behavior Comparison} Figure 7 presents the convergence curves of the Llama 3-8B model with and without the PRADA method during training on GSM8K dataset. The model trained with PRADA (blue curve) demonstrates faster convergence and lower overall loss compared to the vanilla model (orange curve), indicating more efficient learning of domain-agnostic features and better model performance with the PRADA technique. 

\section{Conclusion}

In this work, we propose PRompt-Assisted Domain-Adversarial fine-tuning (PRADA), a novel adversarial fine-tuning framework to enhance the CoT ability of student models in knowledge distillation. Leveraging CoT knowledge distillation, prompt-assisted learning, and domain adversarial optimization, PRADA mitigates the degradation of generalization performance during Chain-of-Thought (CoT) reasoning distillation. This research explores factors causing CoT degradation, designs novel adversarial optimization techniques to improve the fidelity of distilled reasoning while preserving CoT power for portable LLMs. Extensive evaluations across twelve datasets highlight PRADA’s superiority over prior CoT distillation methods, especially in scenarios requiring robust cross-domain generalization.

\bibliographystyle{named}
\bibliography{ijcai25}

\end{document}